\documentclass[11pt]{article}
\usepackage{silence}
\usepackage[final]{acl}

\usepackage{natbib}
\usepackage{tabularx}
\usepackage{times}
\usepackage{latexsym}
\usepackage[T1]{fontenc}
\usepackage[utf8]{inputenc}
\usepackage{microtype}
\usepackage{graphicx}
\usepackage{booktabs}
\usepackage{amsmath}
\usepackage{amssymb}
\usepackage{array}
\usepackage{multirow}
\usepackage{xcolor}
\usepackage{float}
\usepackage{placeins}
\usepackage{tikz}
\usetikzlibrary{shapes.geometric, arrows.meta, positioning, fit, backgrounds, calc}
\title{BERT-based Models vs. Large Language Models for Low-Resource Named Entity Recognition: A Comparative Study on Marathi}

\author{
Hariom Ingle$^{1,3}$ \quad Ronit Ghode$^{1,3}$ \quad Ishwari Gondkar$^{1,3}$ \quad Jidnyasa Harad$^{1,3}$ \quad Raviraj Joshi$^{2,3}$ \\[4pt]
$^{1}$Department of Information Technology, PICT, Pune, India \\
$^{2}$Indian Institute of Technology Madras, Chennai, India \\
$^{3}$L3Cube Labs, Pune, India \\
\vspace{3pt}
\small\texttt{\url{ravirajoshi@gmail.com}}
}

\begin{document}
\maketitle

\begin{abstract}
Named Entity Recognition (NER) for low-resource languages such as Marathi remains a challenging task due to limited annotated resources and linguistic complexity. Although recent Large Language Models (LLMs) have demonstrated strong performance across a wide range of natural language processing tasks, their effectiveness for language-specific NER in low-resource settings remains uncertain. In this study, we fine-tune MahaBERT-v2 on different variants of the MahaNER dataset and systematically compare the performance of these models with an existing MahaNER baseline and prominent general-purpose LLMs, including Gemini, LLaMA-3.3-70B, and Gemma models. All models are evaluated on a Marathi NER test dataset using standard metrics of precision, recall, and F1-score. The experimental results show that the fine-tuned MahaBERT-based models consistently outperform both the baseline and all evaluated LLMs, with the fine-tuned models achieving F1-scores ranging from 0.88 to 0.91, surpassing the existing MahaNER model (0.8843) and significantly exceeding the performance of LLM-based approaches, whose F1-scores range from 0.57 to 0.69. These findings demonstrate that task-specific, language-focused models trained on domain-relevant data remain more effective than general-purpose LLMs for Marathi NER, highlighting the continued importance of specialized architectures for low-resource language processing.
\end{abstract}


\section{Introduction}
\label{sec:intro}

Named Entity Recognition (NER) is a fundamental task in Natural Language Processing (NLP) that involves identifying and classifying named entities such as persons, organizations, locations, and other predefined categories within text \cite{li2022survey}. NER serves as a critical component in downstream applications including information extraction, question answering, machine translation, and knowledge graph construction. While significant advancements have been achieved for high-resource languages like English, developing robust NER systems for low-resource languages remains a persistent challenge.

Marathi is an Indo-Aryan language spoken by over 83 million people, primarily in the state of Maharashtra, India. It is one of the major scheduled languages of India and holds official language status. Despite its large speaker base, Marathi is considered a low-resource language in the context of NLP \cite{joshi2022l3cube_mahanlp}. This is because it has limited annotated corpora, few pretrained language models, and relatively little NLP tooling compared to languages like English, French, or even Hindi. Marathi also presents unique linguistic complexities, including rich morphology, inflectional variations, and flexible word order, all of which make tasks like NER harder to solve.

NER in Marathi is particularly difficult for several interconnected reasons. First, Marathi exhibits rich morphology, where a single word can take many different forms depending on its grammatical role, which increases vocabulary size and makes it harder for models to generalise. Second, unlike English, Marathi written in the Devanagari script does not use capital letters, removing a strong signal that English NER models rely on to detect proper nouns \cite{sang2003conll}. Third, the flexible word order of Marathi allows the subject, verb, and object to appear in varying positions within a sentence, making positional features less reliable for entity detection. Fourth, Marathi text, especially in digital and social media, frequently mixes words from Hindi and English, introducing additional ambiguity through code-mixing. Finally, there are very few large, manually annotated Marathi NER datasets available, which limits the training of data-hungry deep learning models and compounds all of the above challenges.

In this work, we address these challenges by fine-tuning variants of the L3Cube-MahaBERT v2 model \cite{joshi2022l3cube} on the MahaNER dataset \cite{patil2022mahaner} under different training configurations. We specifically study the effect of context length expansion using two data augmentation strategies: self-concatenation and random concatenation. In the self-concatenation approach, each training sentence is repeated multiple times ($4\times$ or $10\times$) within a single input sequence. In the random concatenation approach, multiple different sentences are combined into a single longer sequence. We call the resulting model variants MahaNER-normal-repeat-4x, MahaNER-normal-repeat-10x, MahaNER-random-repeat-4x, and MahaNER-random-repeat-10x.

We then compare these fine-tuned models against several general-purpose Large Language Models evaluated in a zero-shot setting, including Gemini, LLaMA-3.3-70B \cite{touvron2024llama}, and Gemma \cite{gemmateam2024gemma}. The goal is to understand whether task-specific fine-tuned models still hold an advantage over powerful general-purpose LLMs for Marathi NER, and what role data augmentation plays in improving NER performance.

The main contributions of this paper are:
\begin{itemize}
\item We fine-tune and evaluate multiple MahaBERT-v2-based NER models using different data augmentation strategies.
\item We compare fine-tuned Marathi NER models against state-of-the-art LLMs in a zero-shot setting.
\item We show that language-specific fine-tuned models significantly outperform general-purpose LLMs for Marathi NER.
\item We provide a detailed analysis of how training data augmentation affects model performance on different test sets.
\end{itemize}

\section{Related Work}
\label{sec:related}

\subsection{Evolution of NER Systems}

NER has a long research history spanning rule-based, statistical, and neural approaches \cite{li2022survey}. Early NER systems used hand-written rules and dictionaries. These systems required a lot of manual effort and did not generalise well to new domains or languages. Statistical models like Hidden Markov Models (HMMs) and Maximum Entropy classifiers then replaced rule-based systems by learning patterns from annotated data. Conditional Random Fields (CRFs) further improved results by modelling dependencies between output labels, making them well-suited for sequence labelling tasks like NER \cite{lample2016neural}. Benchmark datasets such as CoNLL-2003 \cite{sang2003conll} played a central role in driving progress across these approaches.

The rise of deep learning brought significant improvements. Convolutional Neural Networks (CNNs) and Long Short-Term Memory (LSTM) networks, especially Bidirectional LSTMs (BiLSTMs), were shown to learn useful representations of text directly from data. These models reduced the need for hand-crafted features. Combining BiLSTMs with CRF decoding layers became a popular and strong approach for NER \cite{lample2016neural}.

The introduction of transformer-based models \cite{vaswani2017attention}, starting with BERT \cite{devlin2019bert}, fundamentally changed the field. BERT learns deep contextual representations of words by pretraining on large text corpora using masked language modelling. Fine-tuning BERT on NER datasets consistently achieved state-of-the-art results across many languages and benchmarks. Subsequent models such as RoBERTa \cite{liu2019roberta} and T5 \cite{raffel2020t5} further advanced pretraining methodology, demonstrating that better training objectives and more data consistently improve downstream task performance including NER.

\subsection{Marathi NLP Resources}

Marathi NLP has seen growing interest over the past few years. The L3Cube-MahaCorpus provided a large Marathi text corpus, and MahaBERT was pretrained on this corpus to create the first Marathi-specific BERT model \cite{joshi2022l3cube}. This model significantly outperformed multilingual BERT (mBERT) on downstream Marathi tasks.

MahaNER \cite{patil2022mahaner} provided the first large, manually annotated Marathi NER dataset. It defined entity categories including Person, Organisation, and Location, and showed that MahaBERT-based models outperform multilingual alternatives like mBERT and XLM-R \cite{conneau2020unsupervised} by a clear margin. Later work introduced social media NER for Marathi through the L3Cube-MahaSocialNER dataset \cite{chaudhari2024mahsocialner}, which highlighted the challenges of informal text and code-mixing in low-resource settings.

\subsection{LLMs for NER}

Large Language Models such as GPT-4, LLaMA \cite{touvron2024llama}, and Gemma \cite{gemmateam2024gemma} have shown impressive performance on many NLP tasks in a zero-shot or few-shot setting \cite{brown2020fewshot}. Chain-of-thought prompting \cite{wei2022chain} has further improved LLM reasoning on complex tasks. However, for structured prediction tasks like NER, which require precise token-level labelling, LLMs often struggle. They may produce inconsistent output formats, miss entity boundaries, or fail to follow labelling instructions consistently. This problem is more pronounced for low-resource languages where the LLM's pretraining data is sparse. Despite these known limitations, a systematic comparison between fine-tuned Marathi NER models and state-of-the-art LLMs has not been done before. Our work fills this gap.

\section{Dataset}
\label{sec:dataset}

\subsection{MahaNER Corpus}

All experiments in this paper are based on the MahaNER corpus \cite{patil2022mahaner}, which is the primary benchmark dataset for Marathi Named Entity Recognition. MahaNER was constructed by manually annotating Marathi news text with entity labels following annotation conventions similar to CoNLL-2003 \cite{sang2003conll}. The dataset covers standard NER entity types including Person (PER), Organisation (ORG), and Location (LOC), along with a few other categories. It is the largest and most widely used annotated Marathi NER resource available.

The corpus was annotated using a non-IOB tagging scheme, where each token is assigned one of the following label types:
\begin{itemize}
    \item \textbf{TYPE:} Token is a named entity of a given type (e.g., PER, ORG, LOC).
    \item \textbf{O:} Token is outside any named entity.
\end{itemize}

Table~\ref{tab:mahanerstat} shows the approximate size and entity distribution of the MahaNER corpus.

\begin{table}[h!]
\centering
\small
\renewcommand{\arraystretch}{1.2}
\begin{tabular}{@{}lr@{}}
\toprule
\textbf{Property} & \textbf{Value} \\
\midrule
Total sentences      & $\sim$25,000 \\
Total tokens         & $\sim$400,000 \\
Entity types         & PER, ORG, LOC, MISC \\
Annotation scheme    & Non-IOB \\
Language             & Marathi (Devanagari) \\
Domain               & News text \\
\bottomrule
\end{tabular}
\caption{Key properties of the MahaNER corpus used in this study.}
\label{tab:mahanerstat}
\end{table}

\subsection{Test Dataset Variants}

Three test dataset variants were derived from the MahaNER corpus for evaluation. Table~\ref{tab:datasets} summarises each variant along with its construction strategy and objective.

\begin{table}[h!]
\centering
\small
\renewcommand{\arraystretch}{1.2}
\begin{tabular}{@{}lll@{}}
\toprule
\textbf{Dataset}  & \textbf{Objective} \\
\midrule
Normal              & Baseline \\
Normal Repeat 4x  & Consistency \\
Random Repeat 4x    & Context diversity \\
\bottomrule
\end{tabular}
\caption{MahaNER-derived test datasets. Each variant targets a different aspect of model robustness.}
\label{tab:datasets}
\end{table}

The \textbf{Normal} dataset is the standard test split of MahaNER with naturally occurring entity distributions and no modifications. It serves as the primary baseline for comparison.

The \textbf{Normal Repeat 4x} dataset is created by repeating each test sentence four times within a single sequence. This tests whether models can make consistent predictions when the same sentence appears multiple times in a longer input.

The \textbf{Random Repeat 4x} dataset is built by randomly combining four different test sentences into a single input sequence. This introduces more varied entity patterns within each sequence and tests model robustness to diverse contexts.

\subsection{Training Data Augmentation}

To study the effect of training data augmentation, multiple augmented training sets were created from the MahaNER training split using the same repetition strategies described above. Repetition factors of $4\times$ and $10\times$ were used for both normal-repeat and random-repeat strategies. These augmented training sets give the model exposure to longer input sequences during training, which may improve performance on longer test sequences. Importantly, the augmentation was applied only to the training data; all test sets remain unmodified and in their original form.

\section{Methodology}
\label{sec:methodology}

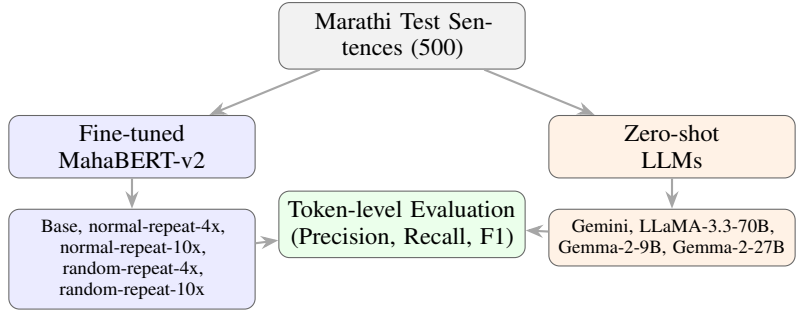
\begin{figure*}[h!]
\centering
\begin{tikzpicture}[
    node distance=0.5cm and 0.4cm,
    box/.style={rectangle, rounded corners=4pt, draw=black!60,
                fill=blue!8, text width=3.0cm, minimum height=0.75cm,
                align=center, font=\small},
    llmbox/.style={rectangle, rounded corners=4pt, draw=black!60,
                fill=orange!10, text width=3.0cm, minimum height=0.75cm,
                align=center, font=\small},
    arrow/.style={-Stealth, thick, draw=gray!70}
]
\node[box, fill=gray!10] (input) {Marathi Test Sentences (500)};

\node[box, below left=0.6cm and 0.3cm of input]  (bert)  {Fine-tuned\\MahaBERT-v2};
\node[llmbox, below right=0.6cm and 0.3cm of input] (llm) {Zero-shot\\LLMs};

\node[box, below=0.4cm of bert, font=\scriptsize, text width=3.0cm] (bmod) {Base, normal-repeat-4x,\\normal-repeat-10x,\\random-repeat-4x, random-repeat-10x};

\node[llmbox, below=0.4cm of llm, font=\scriptsize, text width=3.0cm] (lmod) {Gemini, LLaMA-3.3-70B,\\Gemma-2-9B, Gemma-2-27B};

\node[box, fill=green!8, below=1.6cm of input] (eval) {Token-level Evaluation\\(Precision, Recall, F1)};

\draw[arrow] (input) -- (bert);
\draw[arrow] (input) -- (llm);
\draw[arrow] (bert)  -- (bmod);
\draw[arrow] (llm)   -- (lmod);
\draw[arrow] (bmod)  -- (eval);
\draw[arrow] (lmod)  -- (eval);
\end{tikzpicture}
\caption{Overview of the evaluation pipeline. Both fine-tuned MahaBERT models and zero-shot LLMs are tested on the same 500 Marathi sentences and evaluated using the same metrics.}
\label{fig:overview}
\end{figure*}

This section describes the two types of systems we evaluate: task-specific fine-tuned models and zero-shot Large Language Models. Figure~\ref{fig:overview} gives an overview of the full evaluation pipeline.

\subsection{Task-Specific NER Models}

All task-specific models are based on the L3Cube-MahaBERT v2 architecture \cite{joshi2022l3cube}, a BERT model \cite{devlin2019bert} pretrained on a large Marathi monolingual corpus. Fine-tuning adds a token classification head on top of the pretrained encoder, which assigns a non-IOB NER label to each input token.

Five fine-tuned model variants were trained using different augmented training sets:

\begin{enumerate}
    \item \textbf{MahaNER Base Model:} Trained on the original MahaNER training set without any augmentation.
    \item \textbf{MahaNER-normal-repeat-4x:} Each training sentence is repeated 4 times in one sequence. This increases sequence length and repeats the same context.
    \item \textbf{MahaNER-normal-repeat-10x:} Same as above but with 10 repetitions per sequence, creating much longer inputs.
    \item \textbf{MahaNER-random-repeat-4x:} Four randomly chosen training sentences are combined into one sequence. This exposes the model to diverse contexts within a single training step.
    \item \textbf{MahaNER-random-repeat-10x:} Ten randomly chosen sentences are combined. This provides maximal contextual diversity within each training sequence.
\end{enumerate}

\begin{figure*}[h!]
\centering
\begin{tikzpicture}[
    node distance=0.35cm,
    sent/.style={rectangle, rounded corners=3pt, draw=blue!50,
                 fill=blue!6, minimum width=3.8cm, minimum height=0.55cm,
                 align=center, font=\scriptsize},
    rsent/.style={rectangle, rounded corners=3pt, draw=orange!60,
                 fill=orange!8, minimum width=3.8cm, minimum height=0.55cm,
                 align=center, font=\scriptsize},
    label/.style={font=\small\bfseries}
]
\node[label] (sc) at (0,0) {Self-Concat (SC)};
\node[sent, below=0.2cm of sc] (s1) {Sentence A};
\node[sent, below=of s1]       (s2) {Sentence A};
\node[sent, below=of s2]       (s3) {Sentence A};
\node[sent, below=of s3]       (s4) {Sentence A};
\node[font=\scriptsize, below=0.1cm of s4] {Same sentence $\times$4};

\node[label] at (4.8,0) (rc) {Random-Concat (RC)};
\node[rsent, below=0.2cm of rc] (r1) {Sentence A};
\node[rsent, below=of r1]       (r2) {Sentence B};
\node[rsent, below=of r2]       (r3) {Sentence C};
\node[rsent, below=of r3]       (r4) {Sentence D};
\node[font=\scriptsize, below=0.1cm of r4] {4 different sentences};
\end{tikzpicture}
\caption{Illustration of the two augmentation strategies. Self-concatenation (SC) repeats the same sentence multiple times. Random concatenation (RC) combines different sentences into one longer sequence.}
\label{fig:augmentation}
\end{figure*}
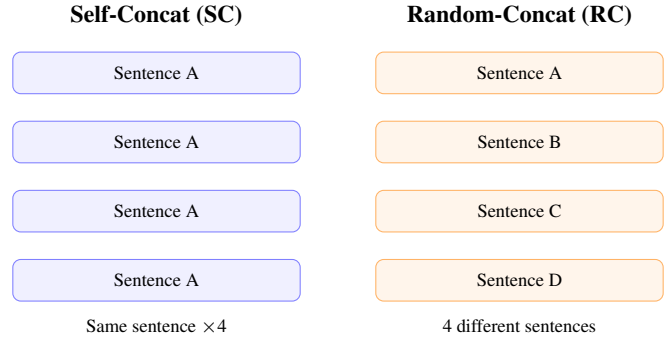

Figure~\ref{fig:augmentation} illustrates the difference between the two augmentation strategies.

All models were fine-tuned using the same hyperparameter settings for fair comparison. A linear classification head was placed on top of the MahaBERT encoder, and training was performed using cross-entropy loss over non-IOB labels.

\subsection{Model Architecture}

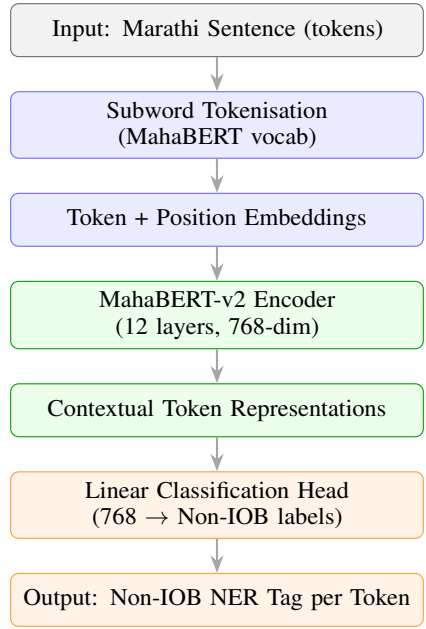
\begin{figure*}[h!]
\centering
\begin{tikzpicture}[
    node distance=0.45cm,
    box/.style={rectangle, rounded corners=3pt, draw=black!60,
                text width=5.2cm, minimum height=0.7cm,
                align=center, font=\small},
    wbox/.style={rectangle, rounded corners=3pt, draw=blue!60, fill=blue!8,
                text width=5.2cm, minimum height=0.7cm,
                align=center, font=\small},
    tbox/.style={rectangle, rounded corners=3pt, draw=green!60!black, fill=green!8,
                text width=5.2cm, minimum height=0.7cm,
                align=center, font=\small},
    obox/.style={rectangle, rounded corners=3pt, draw=orange!80, fill=orange!10,
                text width=5.2cm, minimum height=0.7cm,
                align=center, font=\small},
    arrow/.style={-Stealth, draw=gray!70, thick}
]
\node[box, fill=gray!10]  (input)  {Input: Marathi Sentence (tokens)};
\node[wbox, below=of input]  (tok)    {Subword Tokenisation (MahaBERT vocab)};
\node[wbox, below=of tok]    (emb)    {Token + Position Embeddings};
\node[tbox, below=of emb]    (bert)   {MahaBERT-v2 Encoder (12 layers, 768-dim)};
\node[tbox, below=of bert]   (ctx)    {Contextual Token Representations};
\node[obox, below=of ctx]    (head)   {Linear Classification Head (768 $\to$ Non-IOB labels)};
\node[obox, below=of head]   (out)    {Output: Non-IOB NER Tag per Token};

\foreach \s/\t in {input/tok, tok/emb, emb/bert, bert/ctx, ctx/head, head/out}
    \draw[arrow] (\s) -- (\t);
\end{tikzpicture}
\caption{Architecture of the fine-tuned MahaBERT-v2 NER model. The self-attention mechanism of the transformer encoder \cite{vaswani2017attention} produces contextual representations for each token, which are then passed to a linear classification head that predicts non-IOB NER labels.}
\label{fig:arch}
\end{figure*}

Figure~\ref{fig:arch} shows the architecture of the fine-tuned MahaBERT NER model used in this study.

\subsection{Large Language Model Evaluation}

Four LLMs were evaluated on the same test sentences in a zero-shot setting:
\begin{itemize}
    \item \textbf{Gemini:} Google's general-purpose multimodal LLM.
    \item \textbf{LLaMA-3.3-70B-Versatile} \cite{touvron2024llama}: Meta's open-weight large language model with 70 billion parameters.
    \item \textbf{Gemma-2-9B-IT} \cite{gemmateam2024gemma}: A smaller instruction-tuned model from Google.
    \item \textbf{Gemma-2-27B-IT} \cite{gemmateam2024gemma}: A larger instruction-tuned Gemma model.
\end{itemize}

All LLMs were given the same prompt template for each test sentence, inspired by zero-shot prompting strategies from \citet{brown2020fewshot}. The prompt instructed the model to label each token in the sentence with its NER tag in non-IOB format. No examples were provided in the prompt (zero-shot). Token alignment was enforced to ensure that model outputs could be compared directly against the gold labels using the same evaluation metrics.

\section{Evaluation Protocol}
\label{sec:eval}

\subsection{Metrics}

All models were evaluated using the following standard NER metrics computed at the token level:
\begin{itemize}
    \item \textbf{Precision:} Of all tokens that the model labelled as entities, what fraction were actually entities.
    \item \textbf{Recall:} Of all tokens that were actually entities, what fraction did the model correctly identify.
    \item \textbf{F1-score:} The harmonic mean of precision and recall, giving a single balanced measure of performance.
\end{itemize}

\subsection{Fairness of Comparison}

To ensure a fair comparison between fine-tuned models and LLMs, the following steps were taken:
\begin{itemize}
    \item All models were tested on exactly the same 500 Marathi sentences.
    \item All LLMs received the same prompt template with the same formatting instructions.
    \item Token alignment was enforced for LLM outputs to prevent mismatches between model output tokens and gold-standard tokens.
    \item No fine-tuning, few-shot examples, or additional context was provided to the LLMs.
\end{itemize}

\section{Results}
\label{sec:results}

Tables~\ref{tab:normal}--\ref{tab:random} present the token-level Precision, Recall, and F1-score for all evaluated models across the three benchmark datasets.

\subsection{Results on Normal Dataset}

\begin{table}[h!]
\centering
\small
\setlength{\tabcolsep}{4pt}
\begin{tabularx}{\columnwidth}{lccc}
\toprule
\textbf{Model} & \textbf{P} & \textbf{R} & \textbf{F1} \\
\midrule
Gemini                              & 0.61 & 0.85 & 0.69 \\
LLaMA-3.3-70B                       & 0.56 & 0.81 & 0.64 \\
Gemma-2-9B                          & 0.54 & 0.70 & 0.58 \\
Gemma-2-27B                         & 0.51 & 0.75 & 0.57 \\
\textbf{MahaNER Base Model}         & \textbf{0.90} & \textbf{0.91} & \textbf{0.91} \\
MahaNER-normal-repeat-4x            & 0.88 & 0.92 & 0.90 \\
MahaNER-normal-repeat-10x           & 0.88 & 0.91 & 0.89 \\
MahaNER-random-repeat-4x            & 0.90 & 0.90 & 0.90 \\
MahaNER-random-repeat-10x           & 0.89 & 0.91 & 0.90 \\
\bottomrule
\end{tabularx}
\caption{Performance on the normal (standard) test dataset.}
\label{tab:normal}
\end{table}

On the normal test set, the MahaNER Base Model achieves the highest F1-score of 0.91, closely followed by the MahaNER-normal-repeat-4x, MahaNER-random-repeat-4x, and MahaNER-random-repeat-10x variants at 0.90. The existing MahaNER baseline achieves 0.8843. All fine-tuned models significantly outperform all LLMs. Gemini achieves the best LLM F1 at 0.69, while Gemma-2-27B performs the worst at 0.57.

\subsection{Results on Normal Repeat 4x Dataset}

\begin{table}[h!]
\centering
\small
\setlength{\tabcolsep}{4pt}
\begin{tabularx}{\columnwidth}{lccc}
\toprule
\textbf{Model} & \textbf{P} & \textbf{R} & \textbf{F1} \\
\midrule
Gemini                              & 0.65 & 0.80 & 0.70 \\
LLaMA-3.3-70B                       & 0.50 & 0.77 & 0.58 \\
Gemma-2-9B                          & 0.44 & 0.56 & 0.46 \\
Gemma-2-27B                         & 0.49 & 0.69 & 0.55 \\
MahaNER Base Model                  & 0.91 & 0.89 & 0.90 \\
\textbf{MahaNER-normal-repeat-4x}   & \textbf{0.90} & \textbf{0.93} & \textbf{0.91} \\
MahaNER-normal-repeat-10x           & 0.90 & 0.92 & 0.91 \\
MahaNER-random-repeat-4x            & 0.91 & 0.86 & 0.88 \\
MahaNER-random-repeat-10x           & 0.93 & 0.86 & 0.88 \\
\bottomrule
\end{tabularx}
\caption{Performance on the Normal Repeat 4 test dataset.}
\label{tab:repeated}
\end{table}

On the repeated test set, MahaNER-normal-repeat-4x and MahaNER-normal-repeat-10x both achieve F1 of 0.91. This suggests that training with repeated sequences helps models handle longer, repetitive inputs. The existing MahaNER baseline drops slightly to 0.8668 on this variant, while Gemini's F1 improves marginally to 0.70.

\subsection{Results on Random Repeat 4x Dataset}

\begin{table}[h!]
\centering
\small
\setlength{\tabcolsep}{4pt}
\begin{tabularx}{\columnwidth}{lccc}
\toprule
\textbf{Model} & \textbf{P} & \textbf{R} & \textbf{F1} \\
\midrule
Gemini                              & 0.61 & 0.78 & 0.68 \\
LLaMA-3.3-70B                       & 0.45 & 0.55 & 0.49 \\
Gemma-2-9B                          & 0.35 & 0.37 & 0.35 \\
Gemma-2-27B                         & 0.45 & 0.59 & 0.49 \\
MahaNER Base Model                  & 0.90 & 0.87 & 0.88 \\
MahaNER-normal-repeat-4x            & 0.88 & 0.86 & 0.87 \\
MahaNER-normal-repeat-10x           & 0.89 & 0.86 & 0.87 \\
\textbf{MahaNER-random-repeat-4x}   & \textbf{0.87} & \textbf{0.90} & \textbf{0.89} \\
MahaNER-random-repeat-10x           & 0.87 & 0.90 & 0.88 \\
\bottomrule
\end{tabularx}
\caption{Performance on the Random Repeat 4 test dataset.}
\label{tab:random}
\end{table}

On the random repeat test set, MahaNER-random-repeat-4x achieves the best F1 of 0.89. This is expected, since training with random concatenation better prepares the model for sequences containing diverse entity contexts. LLM performance drops notably on this dataset, with LLaMA and Gemma models falling to F1 scores between 0.35 and 0.49.

\subsection{Summary of Results}

Table~\ref{tab:summary} provides a summary comparison of the best-performing fine-tuned model versus the best LLM across all three test sets.

\begin{table}[h!]
\centering
\small
\renewcommand{\arraystretch}{1.3}
\begin{tabularx}{\columnwidth}{lXcc}
\toprule
\textbf{Test Set} & \textbf{Best Model} & \textbf{F1} & \textbf{LLM F1} \\
\midrule
Normal          & MahaNER Base Model        & 0.91 & 0.69 \\
Normal Repeat 4x  & MahaNER-normal-repeat-4x  & 0.91 & 0.70 \\
Random Repeat 4x  & MahaNER-random-repeat-4x  & 0.89 & 0.68 \\
\bottomrule
\end{tabularx}
\caption{Best fine-tuned model vs best LLM (Gemini) across all test sets.}
\label{tab:summary}
\end{table}

Across all three test sets, fine-tuned MahaBERT models outperform the best LLM (Gemini) by more than 0.20 F1 points. This gap is consistent and large, confirming that task-specific fine-tuning on Marathi data remains far superior to zero-shot LLM inference for Marathi NER.

\section{Discussion}
\label{sec:discussion}

\subsection{Fine-tuned Models vs LLMs}

The experimental results consistently show that fine-tuned MahaBERT models outperform all evaluated LLMs across every test set. The best fine-tuned model achieves an F1-score of 0.91, while the best LLM (Gemini) reaches only 0.69 to 0.70 depending on the test set. This is a gap of more than 20 F1 points, which is very large in NER evaluation terms.

The main reason for this gap is that fine-tuned models have been trained directly on Marathi NER data. They have learned the specific entity patterns, label distributions, and linguistic structures present in Marathi text. LLMs, on the other hand, are general-purpose models that have not been explicitly trained to perform token-level NER in Marathi. Even though LLMs are much larger and trained on more text overall \cite{brown2020fewshot}, they lack the task-specific and language-specific supervision needed for high-quality Marathi NER. This is consistent with prior findings that cross-lingual models underperform language-specific ones on Marathi tasks \cite{conneau2020unsupervised}.

\subsection{Effect of Data Augmentation}

The augmentation experiments reveal a consistent pattern: models trained with augmented data generalise better to the corresponding augmented test sets. MahaNER-normal-repeat-4x and MahaNER-normal-repeat-10x models excel on the Normal Repeat 4x test set, while MahaNER-random-repeat-4x achieves the best score on the Random Repeat 4x test set. This suggests that pretraining-style data diversity strategies \cite{liu2019roberta, raffel2020t5} are also beneficial when applied to fine-tuning data for low-resource NER tasks.

\subsection{LLM Behaviour}

Among the LLMs, Gemini consistently performs the best, with recall values above 0.78 on all test sets. However, Gemini's precision is low (around 0.61), meaning it often labels tokens as entities when they are not. This high-recall, low-precision pattern suggests that Gemini tends to over-predict entities rather than miss them. LLaMA \cite{touvron2024llama} and Gemma \cite{gemmateam2024gemma} models show lower performance overall, especially on the random repeat test set where their F1 scores fall to as low as 0.35.

A key challenge for LLMs in this task is consistent token-level output formatting. Unlike discriminative models which always produce one label per token, generative LLMs may skip tokens, merge them, or produce labels in inconsistent formats. Chain-of-thought prompting \cite{wei2022chain} might help but was not explored in this zero-shot evaluation. These alignment issues directly hurt precision and recall scores.

\section*{Limitations}
\label{sec:limitations}

While this study provides a thorough comparison between fine-tuned models and LLMs, there are several limitations that should be noted:

\begin{itemize}
    \item \textbf{Zero-shot LLM evaluation only:} All LLMs were evaluated in a zero-shot setting. Providing a few labelled examples (few-shot prompting) \cite{brown2020fewshot} or fine-tuning LLMs directly on Marathi NER data might yield better LLM results. This comparison is left for future work.
    \item \textbf{Single domain:} All data comes from Marathi news text \cite{patil2022mahaner}. Models trained and tested on this domain may not generalise well to other domains such as social media \cite{chaudhari2024mahsocialner}, legal text, or conversational speech.
    \item \textbf{Limited augmentation strategies:} The augmentation strategies explored in this paper are limited to sentence repetition and random sentence combination. Other augmentation methods such as back-translation, synonym replacement, or entity substitution were not explored.
    \item \textbf{Fixed prompt template for LLMs:} All LLMs used the same prompt. Different prompt designs or chain-of-thought strategies \cite{wei2022chain} might yield different results. Prompt sensitivity analysis was not conducted.
    \item \textbf{Non-IOB scheme:} The experiments use a non-IOB tagging scheme. Results may differ under other schemes such as IOB2 or BIOES, which are commonly used in benchmarks like CoNLL-2003 \cite{sang2003conll}.
\end{itemize}

\section{Conclusion}
\label{sec:conclusion}

This paper presented a comparative evaluation of fine-tuned MahaBERT-based Marathi NER models against modern general-purpose Large Language Models, including Gemini, LLaMA-3.3-70B \cite{touvron2024llama}, and Gemma variants \cite{gemmateam2024gemma}. We evaluated all systems on three MahaNER-derived test datasets \cite{patil2022mahaner}: a standard benchmark, a repeated-sentence dataset, and a random-combination dataset.

The main findings of this study are:
\begin{itemize}
    \item Fine-tuned MahaBERT models \cite{joshi2022l3cube} consistently and significantly outperform all evaluated LLMs on all three test sets. The best fine-tuned model achieves an F1-score of 0.91, compared to 0.69--0.70 for the best LLM.
    \item Data augmentation through random concatenation (MahaNER-random-repeat-4x/10x) improves model robustness on diverse test inputs.
    \item Self-concatenation augmentation (MahaNER-normal-repeat-4x/10x) helps models generalise better to repeated-sentence test inputs.
    \item LLMs show high recall but low precision for Marathi NER, suggesting they tend to over-predict entity spans.
    \item Language-specific pretraining and supervised fine-tuning remain the most effective approach for low-resource Marathi NER, consistent with findings from multilingual NER research \cite{conneau2020unsupervised, li2022survey}.
\end{itemize}

Future work may explore few-shot or fine-tuned LLM approaches for Marathi NER, additional augmentation strategies, expansion to other Marathi domains such as social media and legal text \cite{chaudhari2024mahsocialner}, and joint entity and relation extraction systems.


\section*{Acknowledgements}

This work was carried out under the mentorship of L3Cube, Pune. We would like to express our gratitude towards our mentor for his continuous support and encouragement. This work is a part of the L3Cube-MahaNLP project \cite{joshi2022l3cube_mahanlp}. 

\bibliography{main}

\end{document}